\documentclass[10pt,twocolumn,letterpaper]{article}
\usepackage{}
\usepackage{iccv}
\usepackage{times}
\usepackage{epsfig}
\usepackage{graphicx}
\usepackage{amsmath}
\usepackage{amssymb}
\usepackage{cases}
\usepackage{txfonts}
\usepackage{multirow}
\usepackage{booktabs}

\newcommand\blfootnote[1]{%
  \begingroup
  \renewcommand\thefootnote{}\footnote{#1}%
  \addtocounter{footnote}{-1}%
  \endgroup
}

\usepackage[pagebackref=true,breaklinks=true,letterpaper=true,colorlinks,bookmarks=false]{hyperref}

\iccvfinalcopy 

\def\iccvPaperID{1776} 

\newcommand{\hua}[1]{{\mathcal #1}}

\ificcvfinal\fi
\begin{document}

\title{Pose-Guided Human Parsing with Deep-Learned Features}
\author{
Fangting Xia, Jun Zhu$^{*}$, Peng Wang$^{*}$, Alan Yuille\\
University of California, Los Angeles
}

\maketitle

\begin{abstract}
   Parsing human body into semantic regions is crucial to human-centric analysis. In this paper, we propose a segment-based parsing pipeline that explores human pose information, i.e.\ the joint location of a human model, which improves the part proposal, accelerates the inference and regularizes the parsing process at the same time. Specifically, we first generate part segment proposals with respect to human joints predicted by a deep model~\cite{Chen14_IDPR}, then part-specific ranking models are trained for segment selection using both pose-based features and deep-learned part potential features. Finally, the best ensemble of the proposed part segments are inferred though an And-Or Graph.
    We evaluate our approach on the popular Penn-Fudan pedestrian parsing dataset~\cite{Wang07_FudanPenn}, and demonstrate the effectiveness of using the pose information for each stage of the parsing pipeline. Finally, we show that our approach yields superior part segmentation accuracy comparing to the state-of-the-art methods.
\end{abstract}
\vspace{-\baselineskip}

\section{Introduction} \label{sec:intro}
\blfootnote{* indicates equal contributions.}

\begin{figure}[!t]
\begin{center}
   \includegraphics[width=0.98\linewidth]{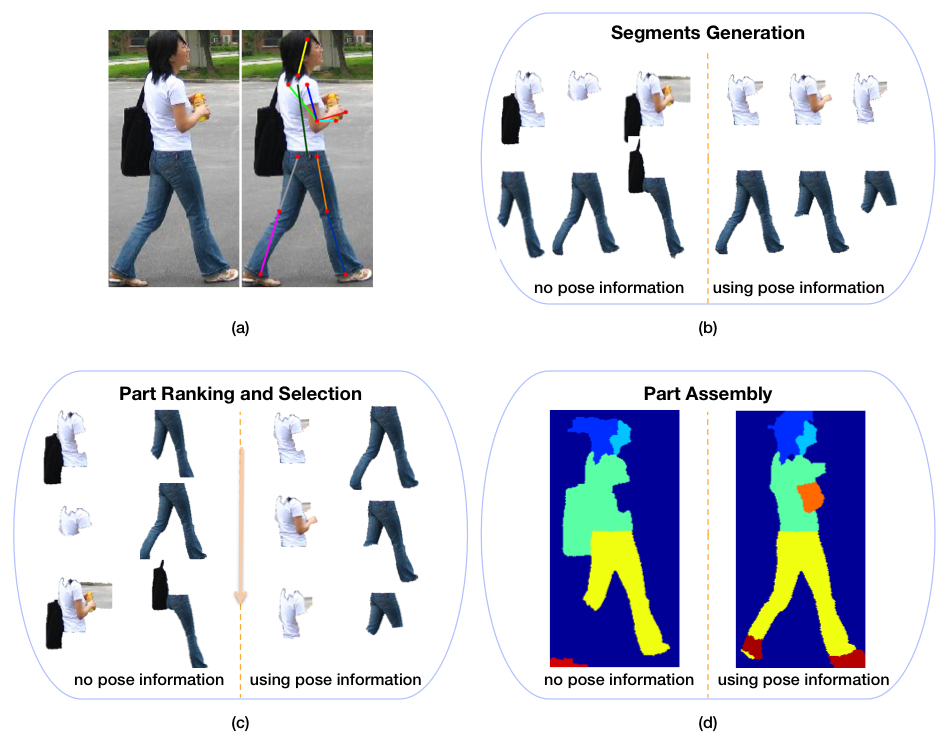}
\end{center}
   \caption{ Human parsing using effective pose cues and deep cues.
   (a) Left: original image. Right: predicted pose.
   (b) Pose-based part proposal. Left panel: without pose information. Right panel: with pose information.
   (c) Part ranking and selection. Left panel: leading part proposals without pose cues. Right panel: leading part proposals with pose cues.
   (d) Final parsing results. Left panel: without pose cues. Right panel: with pose cues.}
\label{fig:motivation}
   \vspace{-\baselineskip}
\end{figure}

The goal of human parsing is to partition the human body into different semantic parts such as hair, head, torso, arms, legs, etc, which provides rich descriptions for human-centric analysis, and thus becomes increasingly important to many computer vision applications, including content-based image/video retrieval~\cite{Weber11_PersonRetrieval,Vaquero09_PeopleSearch}, person re-identification~\cite{Ma11_BodyPrior_ReID,Cheng11_CPS_ReID}, video surveillance~\cite{Vaquero09_PeopleSearch,Yang11_RealTime_ClothRecog_Surveillance,Luo13_PedParsing_deepDecompNet}, action recognition~\cite{Wang12_DHPM_HumanParsing_ActionRecog,Wang13_PoseBasedActionRecog,Zhu13_Acton_ActionRecog} and clothes fashion recognition~\cite{Yamaguchi12_ParsingClothFashion}. However, it is very challenging in real-life scenarios due to variability in human appearances and shapes, caused by large numbers of human poses, clothes types, and occlusion/self-occlusion patterns.

\begin{figure*}[t]
\begin{center}
   \includegraphics[width=0.90\linewidth]{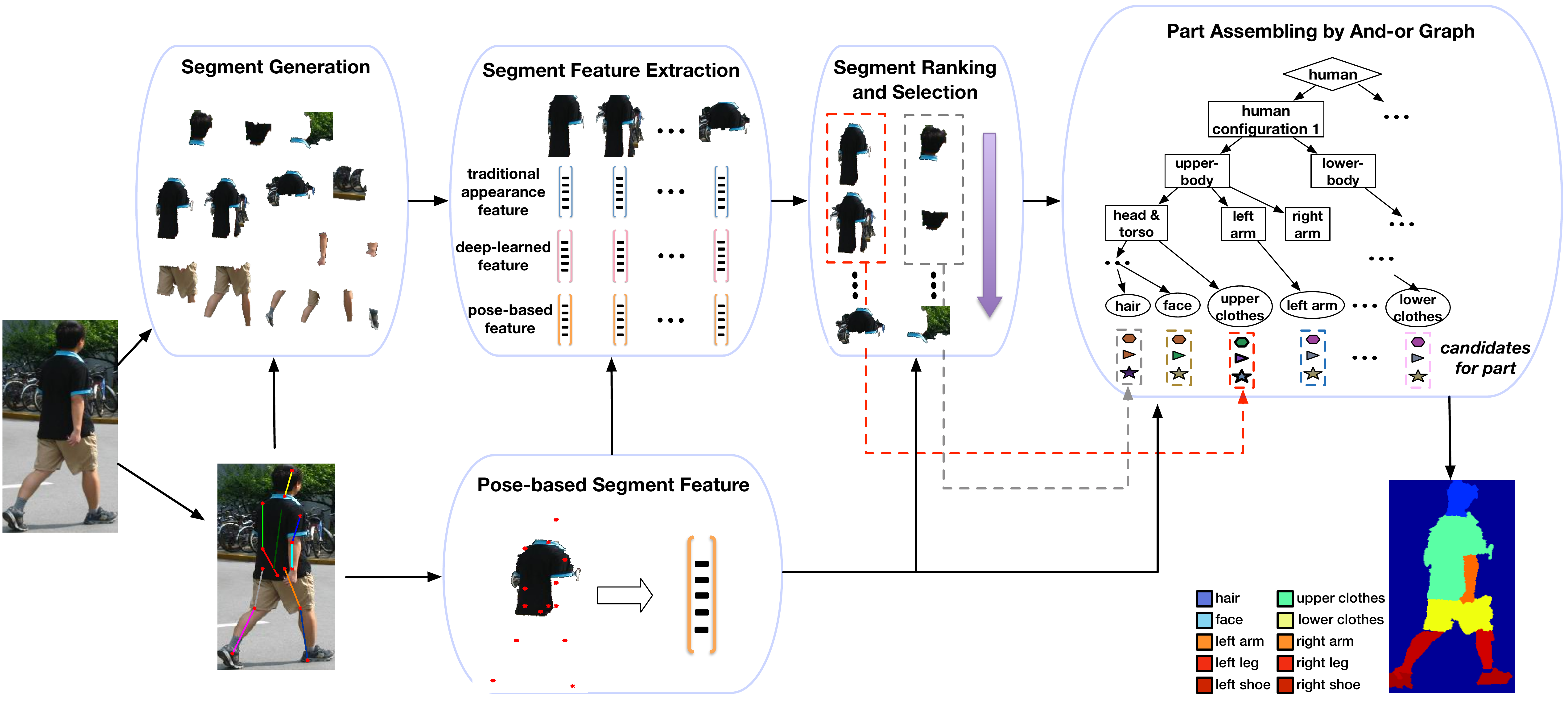}
\end{center}
   \caption{Illustration of our human parsing pipeline.}
\label{fig:framework}
\vspace{-\baselineskip}
\end{figure*}

Current state-of-the-art approach for human parsing is the segment-based graphical model framework by first generating segment/region proposals for parts based on appearance similarity, then selecting and assembling the segments by a graphical model~\cite{Bo11_ShapeBased_PedParsing, Yang14_ClothCoParsing, Dong14_UnifiedParsingPoseEst,LiuCVPR15}.
However, using only bottom-up cues has difficulties in locating good part segments with color ambiguities or noisy regions. Previous strategy~\cite{Dong14_UnifiedParsingPoseEst} integrated pose cues to handle the problem, while the pose model is only used at last stage, where the error made from part proposals is inevitably propagated.
As illustrated in Fig.~\ref{fig:motivation} and Fig.~\ref{fig:framework}, our approach synergies the 
 accurate top-down pose cues in all the parsing process with deep learned features, which largely improves the quality of
the part proposal (in Fig.\ref{fig:motivation}(b)), provides robust feature for ranking (in Fig.\ref{fig:motivation}(c)) and regularizes the graphical ensemble (in Fig.\ref{fig:motivation}(d)).

The framework of our approach is illustrated in Fig.~\ref{fig:framework}. Given an image, firstly, as shown at the bottom, the pose information of the human inside is estimated using a deep pose approach~\cite{Chen14_IDPR}  as the overall parsing guidance in each stage. Our estimation can be very accurate due to the human poses in our parsing task is in a constraint situation, i.e. mostly walking and standing.

At the top row, we show the three stages of our parsing pipeline.  In the first stage, a pool of part segment proposals are generated with the information that parts should appear around pose joints, yielding proposals with high precision and recall. 
In the second stage, rich feature description including pose features and appearance features are proposed to describe each segment proposal, based on which a regressor is trained to re-rank the proposals. 
Specifically, the pose feature captures spatial relationship between a proposed segment and the predicted human pose joints, both  locally and globally.
The appearance feature is computed from both hand-designed features and the deep-learned part potential using the fully convolutional network (FCN)~\cite{long_shelhamer_fcn}, which models a part proposal's shape and appearance.
The two top-down features are complementary and provide robust cues to prune false positive proposals from the background.
After pruning the proposals, a small number of high-quality segment proposals for each part category are selected for the part assembling stage.
In the third stage, an And-Or graph (AOG) \cite{Zhu07_ImageGrammar,Zhu08_AOG_HumanBodyParsing,Zhu12wacv_Tangram,Dong14_UnifiedParsingPoseEst} is applied to optimally assemble the selected proposals of different parts into the final parsing result, in which the pose cue is additionally utilized to measure the pairwise context compatibility. 
We evaluate our method on the popular \emph{Penn-Fudan} \cite{Wang07_FudanPenn} pedestrian parsing benchmark, and show that the pose information effectively improve the performance in every stage of our parsing pipeline, and by incorporating the deep-learned potential features, our approach outperforms other state-of-the-arts in this human parsing task with a significant margin.

In summary, the contributions of this work are in three folds:
\begin{itemize}
\vspace{-0.4\baselineskip}
\setlength\itemsep{1pt}
\item[(1)] We develop a human parsing pipeline that systematically explores the top-down pose information at every stage to regularize the model, yielding strong improvements in parsing efficiency and effectiveness w.r.t. the state-of-the-art methods.
\item[(2)] We effectively incorporate the deep learned features for each part proposal, which provide robust representation of part appearance.  
\item[(3)] We propose a novel pose-based geometric feature that models the spatial relationship of different segments and parts, which is substantially important to part selection and composition.
\end{itemize}
\vspace{-0.4\baselineskip}
\section{Related Work} \label{sec:related_work}
\vspace{-0.4\baselineskip}

In the literature of human parsing, the generation, assembling framework produces the state-of-the-art results.  We will first review these parsing methods w.r.t the two stages.

\emph{Part segment proposal generation.}
Previous works \cite{Bo11_ShapeBased_PedParsing, Yamaguchi12_ParsingClothFashion,  Yang14_ClothCoParsing} usually adopt low-level segment-based proposal. For example, \cite{Yamaguchi12_ParsingClothFashion,Yang14_ClothCoParsing} use uniform appearance superpixels as the elemental proposals of body parts.
Some approaches take higher level cues. Bo and Fowlkes \cite{Bo11_ShapeBased_PedParsing} exploited roughly learned part location priors and part mean shapes information, and derive a number of part segments from the gPb-UCM method \cite{Arbelaez09_UCM} using a constrained region merging method. Dong \textit{et al.} \cite{Dong14_UnifiedParsingPoseEst} employed the Parselets \cite{Dong13_Parselets} for proposal to obtain mid-level part semantic information for proposal. However, either low-level, mid-level or rough location proposals may result in many false positives, misleading the later process.
In our approach, we embed the top-down accurate pose joint cues directly into the efficient bottom-up generation algorithm \cite{Humayun14_rigor} to generate ``pose-guided" proposals, which significantly avoids many false positives and improves the segment quality.


\emph{Part assembling.} Given the generated part segments, an assemble model takes in the selected part segments and outputs the final results, where part unary potentials and relative relationships are leveraged in the model.
Bo \textit{et al.} \cite{Bo11_ShapeBased_PedParsing} developed a compositional model that model human parsing into two different levels of body parts. It uses a series of hard geometric constraints, e.g., face and hair should be adjacent, to model relative part geometry in the inference process.
In \cite{Yamaguchi12_ParsingClothFashion,Yang14_ClothCoParsing}, a conditional random field (CRF) is built on top of the superpixels to label part categories.
In \cite{Dong14_UnifiedParsingPoseEst}, a hybrid parsing model (HPM) is proposed to integrate human part parsing and pose estimation, yielding consistent results in both tasks.
In these works, pose information was used in~\cite{Yamaguchi12_ParsingClothFashion} and \cite{Dong14_UnifiedParsingPoseEst} to improve the results.
However, our method differs and improves from previous works in the following several aspects: (1) Rather than hand crafted features used in~\cite{Dong14_UnifiedParsingPoseEst,Yamaguchi12_ParsingClothFashion}, we use the deep learned features both for pose and appearance, yielding more robust representations. (2) Our pose feature descriptor measures consistency between pose and segments both locally and globally. (3) Most importantly, the pose information is embedded systematically in the whole pipeline including the part segment proposal, part selection and assembling, yielding more efficient inference and more robust human body parsing.

Most recently, some studies also try to adopt deep features in this tasks, yielding impressive results. Luo \textit{et al.} \cite{Luo13_PedParsing_deepDecompNet} proposed a deep decompositional network (DDN) to parse pedestrian images into semantic regions, which uses pixel-wise HOG feature map as the input of their system. Liu \textit{et al.} \cite{LiuCVPR15} adopted non-parametric methods and trained a matching deep network to match the input image region to the retrieved ones.
In \cite{BharathCVPR2015,george_part}, they applied the FCN~\cite{long_shelhamer_fcn} to solve human/object parsing in an end-to-end manner.
Wang \textit{et al.} \cite{peng_part} extended it to a two-channel FCN that jointly tackles object segmentation and part segmentation on some animal classes. However, due to lack of the pose to regularize the parsing model, some false positives can not be effectively avoided. Our work gives a first attempt to embed the pose cue with the deep learned parsing strategies, showing that it is an important complementary information. 
\vspace{-0.5\baselineskip}
\section{The pose-guided human parsing pipeline} \label{sec:framework}
\vspace{-0.5\baselineskip}
Given a pedestrian image $I$, we first adopt the existing state-of-the-art pose estimation approach \cite{Chen14_IDPR} to predict a series of human pose joints $\mathcal{L} = \{l_1, l_2, \cdots, l_{n_l}\}$, where $l_j$ denotes the location of the $j$-th pose joint, and $n_l = 14$ in this paper.
Here we use the same joints as those commonly used in the human pose estimation literature \cite{Yang11_FlexMix,Chen14_IDPR}, i.e. forehead, neck, shoulders, elbows, wrists, hips, knees, and ankles.
Based on the human pose joint cues, our human parsing pipeline has three successive steps: \emph{part segment proposal generation}, \emph{part proposal selection}, and \emph{part assembling}, each of which leverages the pose information as shown in Fig.~\ref{fig:framework}.
We will elaborate on the three steps in the following subsections respectively.

\subsection{Pose-guided part segment proposal generation} \label{ssec:seg_prop}
To generate part segment proposals, we adopt the RIGOR segment proposal method \cite{Humayun14_rigor} which is based on the min-cut algorithm \cite{Carreira12_CPMC} and can efficiently generate segments aligning with object boundaries given user defined initial seeds and cutting thresholds.
In our scenario, we generate the seeds based on the predicted pose joint locations.
Specifically, given the observation that the part segments tend to be surrounding the corresponding pose joints, for each joint $j$ we sample a set of seeds at the $5\times 5$ grid locations over a $40 \times 40$ image patch centered at this joint.
We use $8$ different thresholds, yielding $200$ segment proposals in total for each joint.

Further, we prune out duplicate or highly similar segments and construct a segment proposal pool $\mathcal{S} = \{s_1, s_2, \cdots, s_{N}\}$, where the segments are unique to each other.
In detail, we sequentially add the generated segment proposal only if the intersection over union (IoU) value w.r.t. each existing segment is less than $0.95$.
Finally we generate a pool of around $800$ segment proposals for each image, and use them as candidate part segments in latter two steps.

\subsection{Part proposal selection}  \label{ssec:part_ranking}
Directly feeding a number of segments into the part assembling model leads to very high computational cost due to the existence of pairwise or high-order terms in the AOG model.
Thus, for each part we present a proposal selection step to prune the segments with low probability of being that part class, resulting in much less candidate segments for the part assembling step.

Specially, for each segment proposal $s_i \in \mathcal{S}$, we consider multiple features from a variety of cues on appearance, shapes and poses shown as below:
\begin{itemize}
\vspace{-0.4\baselineskip}
\setlength\itemsep{1pt}
\addtolength{\itemindent}{-0.3cm}
\item $\phi^{o2p}(s_i)$, a second order pooling (O2P) feature~\cite{Carreira12_SOP} for describing appearance cues.
\item $\phi^{skin}(s_i)$, an appearance feature capturing skin color cues. We adopt the method of \cite{Khan10_Skin} to produce a skin potential for each pixel in $s_i$, and $\phi^{skin}(s_i)$ is computed via the second order pooling operation on the skin potential map of $s_i$.
\item $\phi^{pbg}(s_i, \mathcal{L})$, a posed-based geometric (PBG) feature we proposed in this paper, which measures the spatial relationship between the segment $s_i$ and the predicted pose joint configuration $\mathcal{L}$. We will elaborate on this feature in Sec. \ref{ssec:pbg_feat}.
\item $\phi^{c-pbg}(s_i, \hua{L})$, a coded posed-based geometric (C-PBG) feature which is computed as the an encoded version of $\phi^{pbg}(s_i, \mathcal{L})$ using a dictionary.
    It linearizes the feature space of $\phi^{pbg}$, and facilitates to learn the linear regressor later.
    The details will be given in Sec.~\ref{ssec:mixture}. 
\item $\phi^{fcn}(s_i, \hua{H})$, a feature computed from the deep-learned potential maps $\hua{H}$ using FCN~\cite{long_shelhamer_fcn}. It measures the compatibility between the low-level segment image patch and high-level part semantic cues from FCN, and we will introduce the details in Sec.~\ref{ssec:fcn}.
\end{itemize}
Our final feature descriptor of $s_i$ is the concatenation of the aforementioned features, i.e.,
\begin{equation} \label{equ:ranking_score}
\begin{aligned}
\phi(s_i, \mathcal{L}, \hua{H}) = & \, [\phi^{o2p}(s_i), \, \phi^{skin}(s_i), \, \phi^{fcn}(s_i, \hua{H}), \\
& \,\, \phi^{pbg}(s_i, \mathcal{L}), \phi^{c-pbg}(s_i, \mathcal{L})]^\text{T}
\end{aligned}
\end{equation}

On basis of this hybrid feature representation, we train a linear support vector regressor (SVR) \cite{Carreira12_SOP} for each part category.
Let $P$ denote the total number of part categories and $p \in \{1, 2, \cdots, P\}$ denote the index of a part category.
The target variable for training SVR is the IoU value between the segment proposal and ground-truth label map of part category $p$.
The output of SVR model is given by Equ.~\eqref{equ:ranking_score}.
\begin{align} \label{equ:ranking_score}
g^{p}(s_i | \mathcal{L}, \hua{H}) = {\beta_{p}}^{\text{T}} \phi(s_i, \mathcal{L}, \hua{H}),
\vspace{-\baselineskip}
\end{align}
where $\beta_{p}$ is the model parameter of SVR for the $p$-th part category.
Thus, for any part category $p$, we rank the segment proposals in $\mathcal{S}$ based on their SVR scores $\{g^{p}(s_i) \, | \, s_i \in \mathcal{S}\}$.
Finally, we select the top-$n_{p}$ scored segments separately for each part category and combine the selected segment proposals from all part categories to form a new segment pool $\mathcal{\tilde{S}} \subseteq \mathcal{S}$.
In this paper, we set $n_{p} = 10$ such that the number of selected segments in $\tilde{\mathcal{S}}$ is much smaller than $N$.

\subsection{Part assembling with And-Or Graph} \label{ssec:part_assemble}
\begin{figure}[!t]
\begin{center}
   \includegraphics[width=\linewidth]{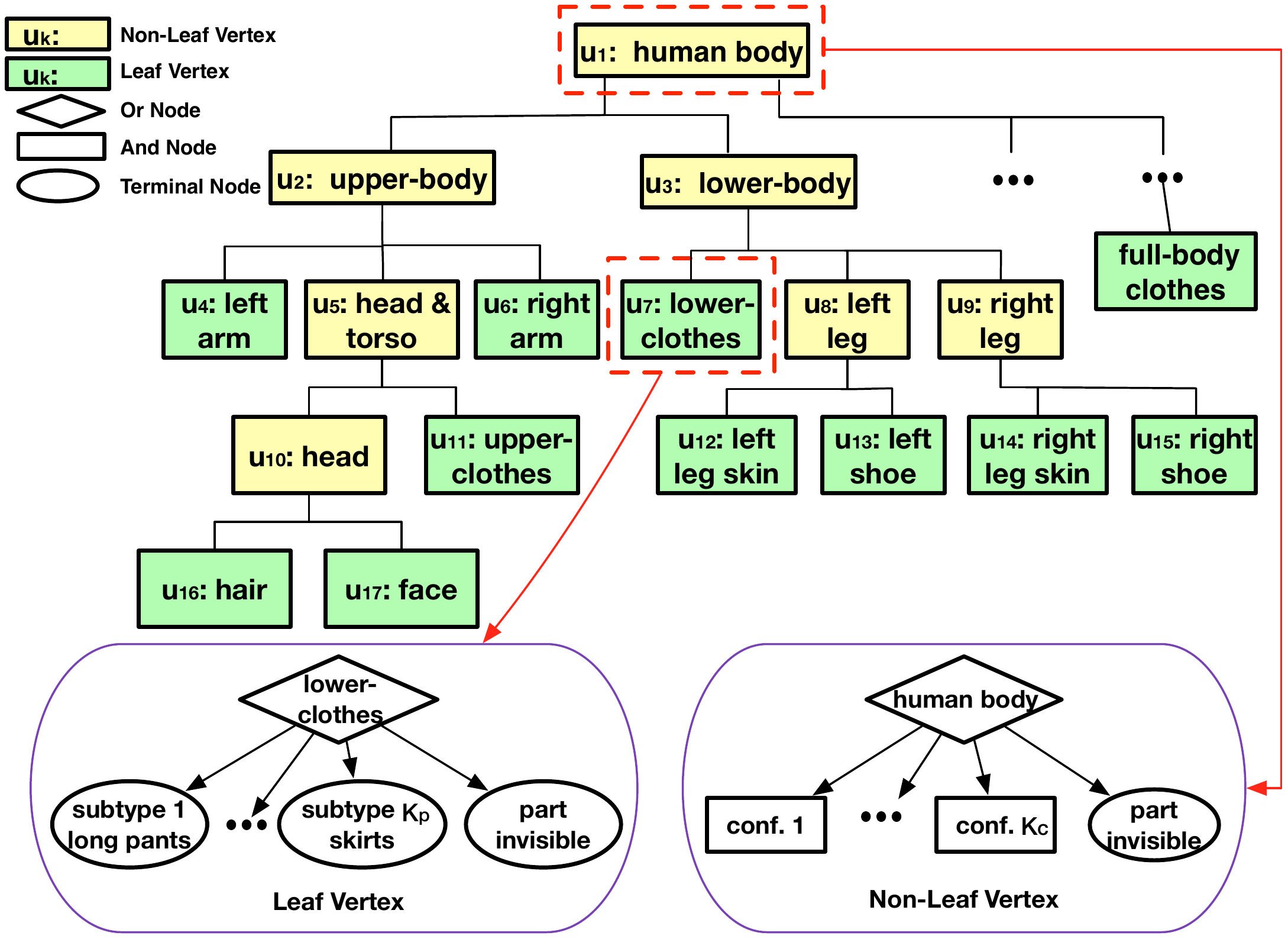}
\end{center}
   \caption{Illustration on the architecture of our AOG model.}
   \vspace{-\baselineskip}
\label{fig:aog}
\end{figure}

There are two different groups of classes (i.e., the \emph{parts} and the \emph{part compositions}) in our AOG model: the part classes are the finest-level constituents of human body, e.g. face, arms, legs, etc.; the part composition classes correspond to intermediate concepts in the hierarchy of semantic human body constituents, e.g. head, upper-body, lower-body, etc., each of which consists of multiple parts.
We list all the parts and part compositions in table \ref{tab:part_def}.
To assemble the selected part segments, we develop a compositional AOG model as illustrated in Fig. \ref{fig:aog}, which facilitates flexible composition structure and standard learning/inference routines.
Let $C$ denote the total number of part compositions.
Formally, our AOG model is defined as a graph $\mathcal{G} = (\mathcal{V}, \mathcal{E})$ where $\mathcal{V} = \mathcal{T} \cup \mathcal{N}$ denotes a set of vertices and $\mathcal{E}$ refers to the set of edges associated.
Meanwhile, $\mathcal{T} = \{ 1, 2, \cdots, P \}$ and $\mathcal{N} = \{ P+1, P+2, \cdots, P+C \}$ denote the set of part indices and the set of part composition indices respectively.
In our AOG, each leaf vertex $p \in \mathcal{T}$ represents one human body part and each non-leaf vertex $c \in \mathcal{N}$ represents one part composition.
The root vertex corresponds to the whole human body while the vertices below correspond to the part compositions or parts at various semantic levels.
Our goal is to parse the human body into a series of part compositions and parts, which is in a hierarchical graph instantiated from the AOG model.

\begin{table} [t!]
\caption{The full list of parts and part compositions in our AOG.}
\begin{center}
\begin{tabular}{c|c}
\hline
\hline
              Part            &   Part Composition \\
\hline
              hair \quad face   &   head \\
              full-body clothes   &   head \& torso \\
              upper-clothes  &   upper-body \\
              left/right arm &   lower-body \\
              lower-clothes   &   left/right leg \\
              left/right leg skin   &   human body \\
              left/right shoe       &    \\
\hline
\hline
\end{tabular}\label{tab:part_def}
\end{center}
\end{table}

The vertex of our AOG is a nested subgraph as illustrated at the bottom in Fig.~\ref{fig:aog}.
For a leaf vertex $p \in \mathcal{T}$, it includes one Or-node followed by a set of terminal nodes as its children.
The terminal nodes correspond to different part types, and the Or-node represents a mixture model indicating the selection of one part type from terminal nodes.
Formally, we define a state variable $z_p \in \{0, 1,2,\cdots,K_p\}$ to indicate that the Or-node selects the $z_p$-th terminal node as the part type for leaf vertex $p$.
As the example of green node in Fig.~\ref{fig:aog}, the lower-clothes part can select one kind of type (e.g. long pants or skirt) from its candidate part types.
Besides there is one special terminal node representing the invisibility of this part due to occlusion/self-occlusion, which corresponds to the state $z_p = 0$.
For a non-leaf vertex $c \in \mathcal{N}$, it includes one Or-node linked by a set of And-nodes plus one terminal node.
The Or-node of non-leaf vertex represents this part composition has several different ways of decompositions into smaller parts and/or part compositions.
The And-node corresponds to one alternative configuration of decomposition for $c$.
As shown in Fig.~\ref{fig:aog}, the non-leaf vertex head can be composed by one of several different configurations of two child vertices (i.e., face and hair).
Similar to the leaf vertices, we also induce a state variable $z_c \in \{0, 1,2,\cdots,K_c\}$ to indicate that the Or-node of part composition $c$ selects the $z_c$-th And-node as the configuration of child vertices for $z_c \neq 0$ or this part composition is invisible when $z_c = 0$.

\begin{figure}[!t]
\begin{center}
   \includegraphics[width=\linewidth]{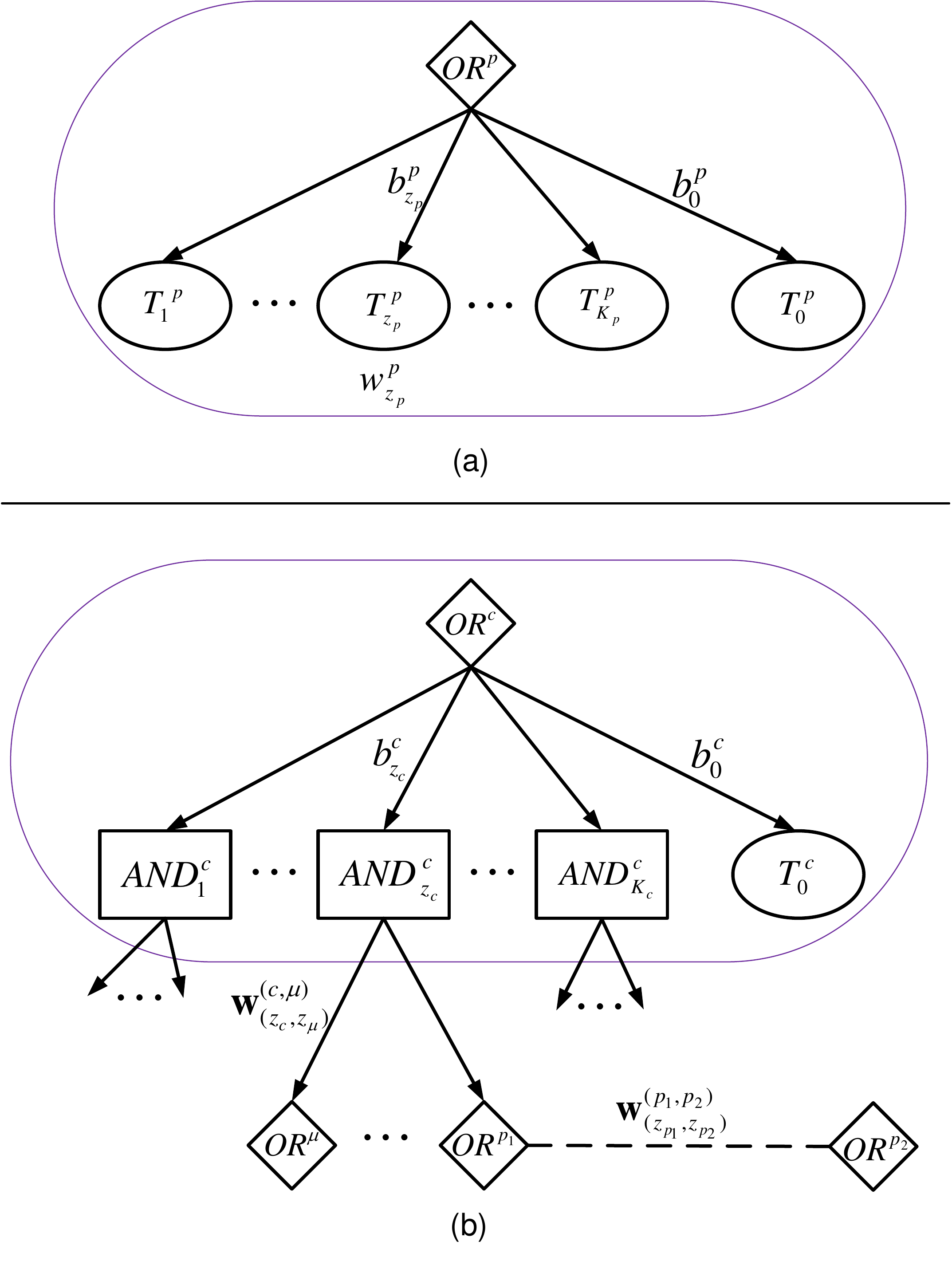}
\end{center}
   \caption{Illustration on the structure of vertices in AOG. (a) leaf vertex; (b) non-leaf vertex. The symbols of \textit{OR}, \textit{AND} and \textit{T} represent the Or-node, And-node and terminal node respectively. Please see Eqn.~\eqref{equ:leaf_score} and Eqn.~\eqref{equ:nonleaf_score} about the notations of model parameters.}
   \vspace{-\baselineskip}
\label{fig:nodes}
\end{figure}

\begin{table*} [t!]
\caption{The list of adjacent part pairs.}
\begin{center}
\begin{tabular}{cc}
\hline
\hline
              Part Composition ($c$) & Adjacent Part Pairs ($\mathcal{R}_{c}$) \\
\hline
              human body      & (upper-clothes, lower-clothes), \, (full-body clothes, left leg skin), \, (full-body clothes, right leg skin)\\
              head                  & (hair, face) \\
              head \& torso     & (upper-clothes, hair), \, (upper-clothes, face), \, (full-body clothes, hair), \, (full-body clothes, face) \\
              upper-body        & (left arm, upper-clothes), \, (right arm, upper-clothes)\\
                                        &  (left arm, full-body clothes), \, (right arm, full-body clothes) \\
              lower-body        & (lower-clothes, left leg skin), \, (lower-clothes, right leg skin) \\
                                        & (lower-clothes, left shoe), \, (lower-clothes, right shoe) \\
              left leg                & (left leg skin, left shoe) \\
              right leg              & (right leg skin, right shoe) \\
\hline
\hline
\end{tabular}\label{tab:rel_part_pair}
\end{center}
\end{table*}


Furthermore, we define another state variable $y$ to indicate the selection of segment from the candidate pool of a part or part composition.
For a leaf vertex $p \in \mathcal{T}$, $y_p \in \{0, 1, 2, \cdots, n_{p,z_p}\}$ represents that the part $p$ selects the $y_p$-th segment proposal (i.e., $s^{p,z_p}_{y_p}$) from the segment pool $\tilde{\mathcal{S}}_{p,z_p}$ output by its segment ranking model on type $z_p$.
Meanwhile, $y_p = 0$ is a special state which coincides with the invisibility pattern of part $p$ (i.e., $z_p = 0$).
To make notation consistent, we use $s^{p,z_p}_{0}$ to represent an "null" segment for part invisibility.
For a non-leaf vertex $c \in \mathcal{N}$, $y_c \in \{0, 1,2,\cdots,n_c\}$ indicates a segment $s^{c,z_c}_{y_c} \in \tilde{\mathcal{S}}_{c,z_c}$ is selected, where $s^{c,z_c}_{y_c}$ is obtained by the union of its child vertices' candidate segments and $\tilde{\mathcal{S}}_{c,z_c}$ denotes the candidate segment pool for the $z_c$ And-node.
When $y_c = 0$, likewise, the $s^{c,z_c}_{0}$ represents a null segment indicating the invisibility pattern of part composition $c$.
Let $Ch(c,z_c)$ denote the set of child vertices for part composition $c$ and configuration $z_c$.
Formally, $s^{c,z_c}_{y_c}$ is defined by Equ.~\eqref{equ:part_comp_seg}.
\begin{align} \label{equ:part_comp_seg}
s^{c,z_c}_{y_c} = \Cup_{\mu \in Ch(c,z_c)} s^{\mu,z_{\mu}}_{y_{\mu}},
\vspace{-\baselineskip}
\end{align}
where $\Cup$ represents a pixel-wise union operation of combing the child vertices' segment masks to generate a new segment.

Let $\mathbf{Y} = (y_{1}, y_{2}, \cdots, y_{P}, y_{P+1}, y_{P+2}, \cdots, y_{P+C})$ and $\mathbf{Z} = (z_{1}, z_{2}, \cdots, z_{P}, z_{P+1}, z_{P+2}, \cdots, z_{P+C})$ denote the structural solution of AOG.
We define a global score function of AOG $F(\mathbf{Y}, \mathbf{Z} \, | \, \tilde{\mathcal{S}}, \mathcal{L}, \hua{H})$\footnote{Here $\tilde{\mathcal{S}} = \bigcup\limits_{p \in \mathcal{T}, \, z_p \neq 0} \tilde{\mathcal{S}}_{p,z_p}$} to measure the compatibility between $(\mathbf{Y}, \mathbf{Z})$ and $(\tilde{\mathcal{S}}, \mathcal{L}, \hua{H})$ for image $I$, which can be calculated as shown in Equ.~\eqref{equ:score_func}.
\begin{align}\label{equ:score_func}
F(\mathbf{Y}, \mathbf{Z} | \tilde{\mathcal{S}}, \mathcal{L}, \hua{H}) &= \sum_{p \in \mathcal{T}} f(y_p, z_p)\\
&+ \sum_{c \in \mathcal{N}} f(y_c, z_c, \{ (y_\mu, z_\mu): \mu \in Ch(c, z_c) \}), \nonumber
\vspace{-0.5\baselineskip}
\end{align}
where $f(y_p, z_p)$ is a local scoring function of leaf vertex $p$ and $f(y_c, z_c, \{ (y_\mu, z_\mu): \mu \in Ch(c, z_c) \})$ denotes a scoring function of the partial graph rooted by non-leaf vertex $c$.
For each leaf vertex $p \in \mathcal{T}$ (i.e., a part), we compute $f(y_p, z_p)$ by Equ.~\eqref{equ:leaf_score}.
\begin{numcases}{\label{equ:leaf_score} f(y_p, z_p)=}
b^{p}_{z_p} + w^{p}_{z_p} \cdot g^{p}_{z_p}(s^{p,z_p}_{y_p} | \mathcal{L}, \hua{H}), & $z_p \neq 0$ \\
b^{p}_{0}, & $z_p = 0$ \nonumber
\vspace{-0.5\baselineskip}
\end{numcases}
where $w^{p}_{z_p}$ and $b^{p}_{z_p}$ denote the weight and bias parameters of unary term for part $p$ respectively.
Particularly, $b^{p}_{0}$ is the bias parameter for the invisibility pattern of $p$.
Besides, $g^{p}_{z_p}$ is dependent on the part type $z_p$, implying the regression models defined in Equ.~\eqref{equ:ranking_score} are trained by different parts and types.
Fig.~\ref{fig:nodes} (a) illustrates the structure of a leaf vertex as well as corresponding model parameters.

For each non-leaf vertex $c \in \mathcal{N}$ (i.e., a part composition), we compute $f(y_c, z_c, \{ (y_{\mu}, z_{\mu}): \mu \in Ch(c, z_c) \})$ by Eqn.~\eqref{equ:nonleaf_score}.
\vspace{2mm}
\begin{equation} \label{equ:nonleaf_score}
f(y_c, z_c, \{ (y_{\mu}, z_{\mu}): \mu \in Ch(c, z_c) \})
\end{equation}
\begin{numcases}{=}
b^{c}_{z_c} + u(y_c, z_c, \{ (y_{\mu}, z_{\mu}): \mu \in Ch(c, z_c) \}), & $z_c \neq 0$ \nonumber \\
b^{c}_{0}, & $z_c = 0$ \nonumber
\end{numcases}
where
\begin{align} \label{equ:nonleaf_score_visible}
&u(y_c, z_c, \{ (y_{\mu}, z_{\mu}): \mu \in Ch(c, z_c) \})\\ \nonumber
= & \,\; h(\{ (y_{\mu}, z_{\mu}): \mu \in Ch(c, z_c) \})\\ \nonumber
&+ \sum\limits_{\mu \, \in \, Ch(c, z_c)} {\mathbf{w}^{(c,\mu)}_{(z_c, z_{\mu})}}^{\text{T}} \varphi(s^{c,z_c}_{y_c}, s^{\mu,z_{\mu}}_{y_{\mu}})\\ \nonumber
&+ \sum\limits_{(p_1,p_2) \in \mathcal{R}_c} {\mathbf{w}^{(p_1,p_2)}_{(z_{p_1}, z_{p_2})}}^{\text{T}} \psi(s^{p_1,z_{p_1}}_{y_{p_1}}, s^{p_2,z_{p_2}}_{y_{p_2}} | \mathcal{L}),
\vspace{-0.5\baselineskip}
\end{align}
and
\begin{align}\label{equ:child_score}
&h(\{ (y_{\mu}, z_{\mu}): \mu \in Ch(c, z_c) \}) = \sum\limits_{\mu \, \in \, Ch(c, z_c) \bigcap \mathcal{T}} f(y_{\mu}, z_{\mu})\\ \nonumber
&+ \sum\limits_{\mu \, \in \, Ch(c, z_c) \bigcap \mathcal{N}} f(y_{\mu}, z_{\mu}, \{ (y_{\nu}, z_{\nu}): \nu \in Ch(\mu, z_{\mu}) \}).
\end{align}
Concretely, Eqn.~\eqref{equ:nonleaf_score} can be divided into four terms:
\begin{itemize}
\item[(1)] the bias term of selecting $z_c$ for the Or-node, i.e. $b^{c}_{z_c}$.
    $b^{c}_{0}$ is the bias parameter when part composition $c$ is invisible (In this case, all the descendant vertices are also invisible and thus the latter three terms are zero).
\item[(2)] the sum of scores of its child vertices for the selected And-node, i.e. $h(\{ (y_{\mu}, z_{\mu}): \mu \in Ch(c, z_c) \})$.
\item[(3)] the sum of parent-child pairwise terms (i.e., \emph{vertical edges}) for measuring the spatial compatibility between the segment of part composition $c$ and the segments of its child vertices, i.e. $\sum\limits_{\mu \, \in \, Ch(c, z_c)} {\mathbf{w}^{(c,\mu)}_{(z_c, z_{\mu})}}^{\text{T}} \varphi(s^{c,z_c}_{y_c}, s^{\mu,z_{\mu}}_{y_{\mu}})$, where $\varphi(s^{c,z_c}_{y_c}, s^{\mu,z_{\mu}}_{y_{\mu}})$ denotes a spatial compatibility feature of segment pair $(s^{c,z_c}_{y_c}, s^{\mu,z_{\mu}}_{y_{\mu}})$ and $\mathbf{w}^{(c,\mu)}_{(z_c, z_{\mu})}$ refers to corresponding weight vector.
    Specifically, $\varphi$ is defined by $[dx; dx^{2}; dy; dy^{2}; ds; ds^{2}]$, in which $dx$, $dy$ represent the spatial displacement between the center locations of two segments while $ds$ is the scale ratio of them.
\item[(4)] the sum of pairwise terms (i.e., \emph{side-way edges}) for measuring the geometric compatibility on all segment pairs specified by an adjacent part-pair set $\mathcal{R}_{c}$, which defines a couple of adjacent part pairs for $c$ (e.g., for the part composition of lower body, we consider lower-clothes and leg skin to be an adjacent part pair)\footnote{We list the adjacent part pairs for each part composition in table \ref{tab:rel_part_pair}.}.
    To avoid double counting in recursive computation of Eqn.~\eqref{equ:nonleaf_score_visible}, $\mathcal{R}_{c}$ only includes the relevant part pairs which have at least one child vertex of $c$.
    This side-way pairwise potential corresponds to $\sum\limits_{(p_1,p_2) \in \mathcal{R}_c} {\mathbf{w}^{(p_1,p_2)}_{(z_{p_1}, z_{p_2})}}^{\text{T}} \psi(s^{p_1,z_{p_1}}_{y_{p_1}}, s^{p_2,z_{p_2}}_{y_{p_2}} | \mathcal{L})$ in Eqn.~\eqref{equ:nonleaf_score_visible}, where $\psi(s^{p_1,z_{p_1}}_{y_{p_1}}, s^{p_2,z_{p_2}}_{y_{p_2}} | \mathcal{L})$ represents a geometric compatibility feature of segment pair $(s^{p_1,z_{p_1}}_{y_{p_1}}, s^{p_2,z_{p_2}}_{y_{p_2}})$ and $\mathbf{w}^{(p_1,p_2)}_{(z_{p_1}, z_{p_2})}$ is corresponding weight vector.
    In this paper, we use the coded version of joint-segment geometric feature for $\psi$, which will be elaborated in Sec. \ref{ssec:mixture}.
\end{itemize}
In Fig.~\ref{fig:nodes} (b), we illustrate the structure of a non-leaf vertex and corresponding model parameters.

According to the hierarchical architecture of our AOG, $F(\mathbf{Y}, \mathbf{Z} | \mathcal{L}, \tilde{\mathcal{S}}, \hua{H})$ can be recursively computed by Eqn.~\eqref{equ:leaf_score} and Eqn.~\eqref{equ:nonleaf_score} in a bottom-up manner.
In practice, we first calculate the score for each leaf vertex, and then calculate the scores of non-leaf vertices from the lowest levels to the root part composition vertex (i.e., human).
It is noted that our AOG is not a directed acyclic graph due to the existence of side-way edges, which makes loops among a clique of some vertices.
This prevent our model from using common dynamic programming for inference.
In this paper, we propose a greedy pruning algorithm for model inference and will elaborate on it in Sec. \ref{ssec:inference}.

\vspace{-0.5\baselineskip}
\section{Feature design} \label{sec:feats}
In this section, we elaborate on the aforementioned features in part proposal selection (Sec.~\ref{ssec:part_ranking}) and part assembling (Sec.~\ref{ssec:part_assemble}) steps.

\subsection{The pose-based geometric (PBG) feature} \label{ssec:pbg_feat}
\begin{figure}[!t]
\begin{center}
   \includegraphics[width=0.95\linewidth]{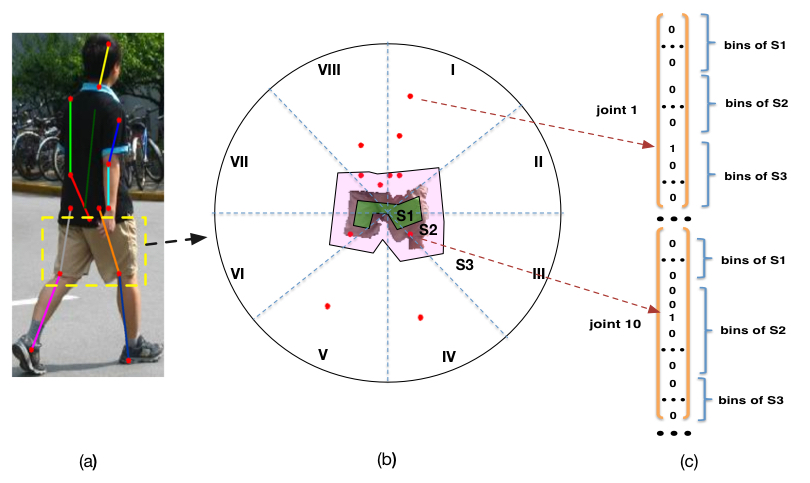}
\end{center}
\caption{Illustration of the pose-based geometric feature (PBG).}
\vspace{-1\baselineskip}
\label{fig:pose_feat}
\end{figure}

\vspace{-0.5\baselineskip}
The PBG feature of a segment $\phi^{pbg}(s_i, \mathcal{L})$ is calculated based on the spatial relationship between the segment $s_i$ and the predicted pose joints $\mathcal{L}$.
As shown in Fig. \ref{fig:pose_feat}, centered at $s_i$, the image is equally divided into eight orientations ($\text{I} - \text{VIII}$) and three region scales ($\text{S1}$, $\text{S2}$ and $\text{S3}$), yielding $24$ spatial bins in total.
Then each joint $l_j \in \mathcal{L}$ falls into one of these spatial bins and produces a $24$ dimensional binary feature, quantizing the spatial relationship of $l_j$ w.r.t. $s_i$.
After that, we concatenate the binary features of all joints, and produce a $24\times 14 = 336$ dimensional binary feature vector to describe the spatial relationship of segment $s_i$ w.r.t. the pose joint configuration $\mathcal{L}$.

Specifically, $\text{S1}$ and $\text{S2}$ are the regions eroded and dilated by 10 pixels from the segments boundary respectively. $\text{S3}$ is the rest region of image.
This segment-dependent definition of region scales depicts meaningful geometric cues from the predicted pose information.
Intuitively, $\text{S1}$ constrains the involved joints inside the segment, while $\text{S2}$ implies the joints tend to be around the segment boundary.
$\text{S3}$ indicates that the joints should be totally outside the segment.
Thus, with our PBG feature, we can learn a more discriminative model by leveraging the geometric compatibility between the segments and pose joints.

\subsection{The coded PBG feature (C-PBG)} \label{ssec:mixture}
The proposed pose feature can be highly non-linear in the feature space which might be suboptimal for linear classification.
This motivates us to encode the pose feature though feature coding to achieve linearity, which has long been proofed to be effective in many vision tasks such as image classification~\cite{DBLP:conf/cvpr/LazebnikSP06, yang2009linear, wang2010locality, Liu2011_LSAQ}, semantic segmentation~\cite{Carreira12_SOP} etc.
In this paper, we adopt a simple soft-assignment quantization (SAQ) coding method \cite{Liu2011_LSAQ} to encode the PBG feature aforementioned.

\begin{figure}[!t]
\begin{center}
   \includegraphics[width=1.02\linewidth]{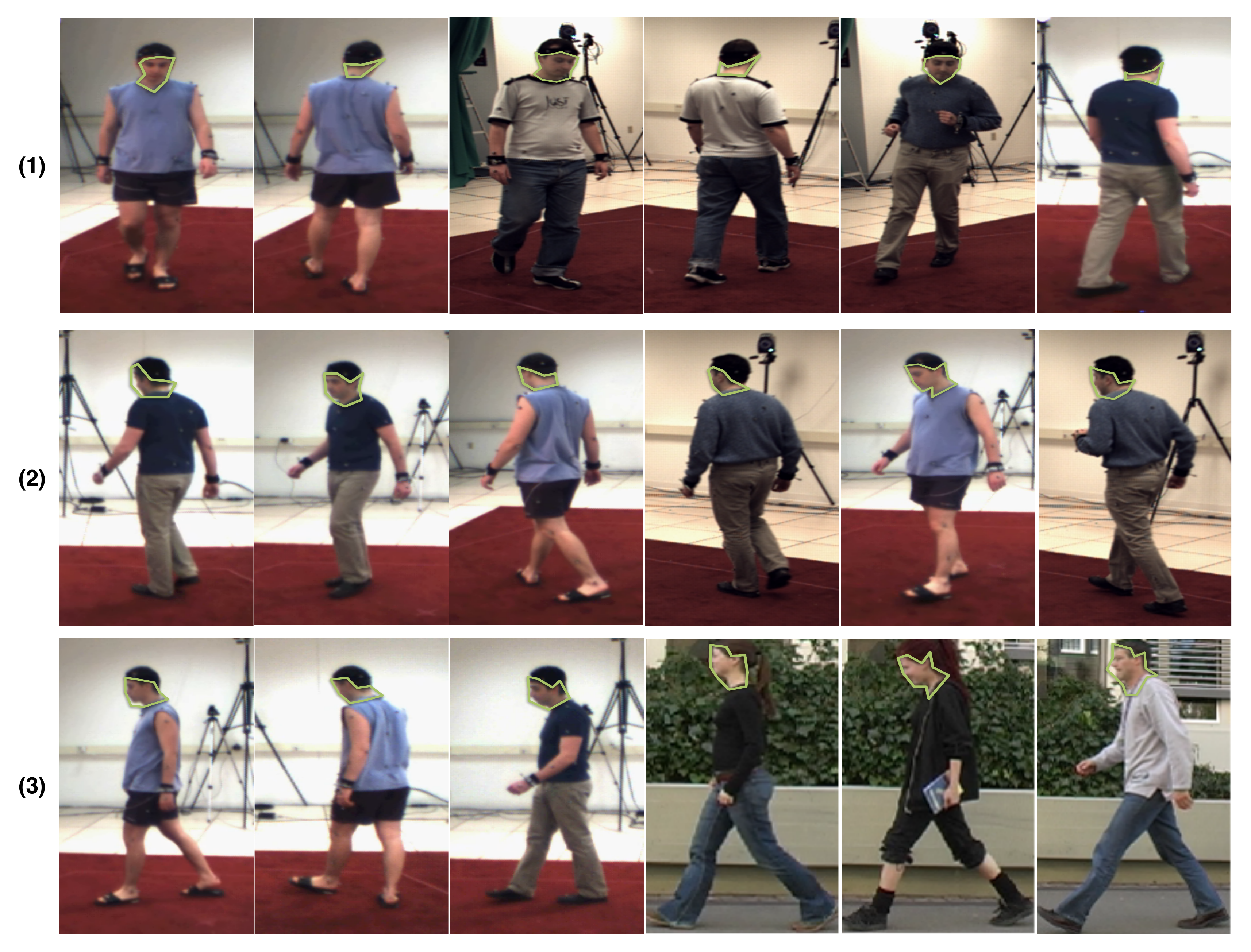}
\end{center}
   \caption{The learned prototypes/clusters for part category face. We show exemplar images for 3 out of 6 clusters. Cluster (1): frontal face or back face. Cluster (2): frontal/back face on the left. Cluster (3): side face on the left. The other clusters correspond to the symmetric patterns w.r.t. those shown here.}
   \vspace{-1\baselineskip}
\label{fig:pose_cluster}
\end{figure}

At first, we collect the ground-truth segments from training images and compute their PBG features.
After that, a dictionary of pose-guided part prototypes $\mathcal{D} = \{ \mathbf{b}_{m} \}_{m=1}^{N_{\mathcal{D}}}$ can be learned via K-means clustering algorithm on the PBG feature representation of segment examples.
To balance different part categories, we separately perform clustering and obtain $N_{p} = 6$ for each part category, resulting in a dictionary of $N_{\mathcal{D}} = K \times N_{p}$ codewords.
Given $\mathcal{D}$, we compute the Euclid distance between original PBG feature of $s_i$ and each prototype $\mathbf{b}_{m}$: $d_{i,m} = \parallel \phi^{pbg}(s_i, \mathcal{L}) - \mathbf{b}_{m} \parallel$.
Thus the coded posed-based geometric (C-PBG) feature is formally defined as the concatenation of both the normalized and un-normalized codes w.r.t. $\mathcal{D}$:
\begin{align} \label{equ:c-pbg}
\phi^{c-pbg}(s_i,\mathcal{L} \, | \, \mathcal{D}) = [a_{i,1}, \cdots, a_{i,N_{\mathcal{D}}}, a^{'}_{i,1}, \cdots, a^{'}_{i,N_{\mathcal{D}}}]^{\text{T}},
\end{align}
where $a_{i,m} = \exp^{(- \lambda d_{i,m})}$ and $a^{'}_{i,m} = \frac{a_{i,m}}{\sum_{j=1}^{N_{\mathcal{D}}} a_{i,j}}$ denote the un-normalized and normalized code values w.r.t. $\mathbf{b}_{m}$ respectively.
$\lambda$ is a hyper-parameter of our coding method.
As introduced in Sec. \ref{ssec:part_ranking}, the C-PBG feature is adopted in training the SVR models for part proposal selection.
Besides, it is used for generating candidate part types in part assembling step.
As illustrated in Fig.~\ref{fig:pose_cluster}, the learned prototypes/clusters generally correspond to different typical views of the face part category.

\begin{figure}[!t]
\begin{center}
   \includegraphics[width=1.02\linewidth]{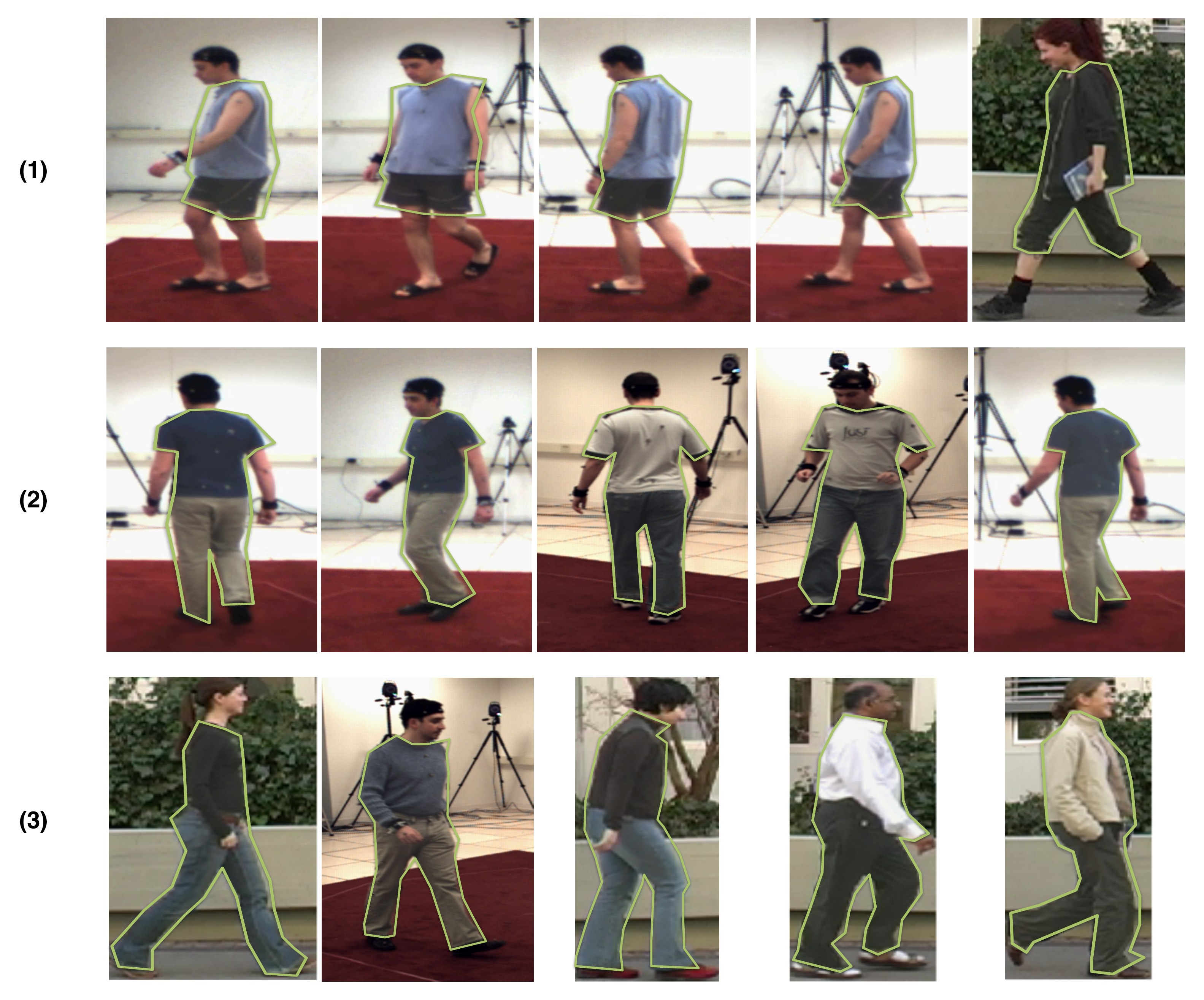}
\end{center}
   \caption{The learned prototypes/clusters for the adjacent part pair \textit{upper-clothes and lower-clothes}. We show 3 out of 8 clusters. Cluster (1): the person with short sleeved upper-clothes and short pants. Cluster (2): the person with short sleeved upper-clothes and long pants. Cluster (3): the person with long sleeved upper-clothes and long pants.}
   \vspace{-1\baselineskip}
\label{fig:pose_pair_cluster}
\end{figure}

In addition, we propose to code the pairwise PBG feature, i.e. $\psi(s^{p_1,z_{p_1}}_{y_{p_1}},s^{p_2,z_{p_2}}_{y_{p_2}} \, | \, \mathcal{L})$ applied in Eqn.~\eqref{equ:nonleaf_score}, for describing the geometric relationship between two adjacent parts $p_1$ and $p_2$.
Specifically, we adopt the same coding process as before but using the concatenated PBG features of a pair of candidate segments $(s^{p_1,z_{p_1}}_{y_{p_1}}, s^{p_2,z_{p_2}}_{y_{p_2}})$.
We perform clustering separately for each adjacent part pair and learn a class-specific dictionary for this pairwise C-PBG feature.
In this paper, the dictionary size is set by $N_{pp} = 8$ for each part pair.
As visualized in Fig.~\ref{fig:pose_pair_cluster}, the learned part-pair prototypes are very meaningful which capture typical view point and part type co-occurrence patterns for adjacent parts.

\subsection{Deep-learned potential} \label{ssec:fcn}
As applied in \cite{BharathCVPR2015, peng_part, george_part}, deep potentials for part, although lack of ability of discovering small parts and global pose regularization, provide strong high-level semantic evidence of part class and shape cues.
Thus, we also combine this potential into the feature representation of segment proposals.
Specifically, we train a FCN with 16s as in~\cite{peng_part} with the output variables to be the part ground truth map.
Given an image, it produces $P+1$\footnote{The additional 1 corresponds to the background class.} potential maps $\hua{H}$, from which we also obtain $P+1$ binary part class label masks $\hua{B}$ though MAP over each potential map.
Thus, for a segment $s_i$, the deep feature $\phi^{fcn}(s_i,\hua{H})$ consists of three components:
(1) the mean value inside $s_i$ of all maps in $\hua{H}$;
(2) the mean value along the contour of $s_i$ from the maps in $\hua{H}$;
(3) The IoU value between $s_i$ to all the $K+1$ masks in $\hua{B}$, i.e. $IoU(s_i, \hua{B})$.

 \vspace{-0.5\baselineskip}
\section{Learning and inference for AOG} \label{sec:learn_infer}
   \vspace{-0.5\baselineskip}
In this section, we first introduce the learning algorithm on our AOG model, and then elaborate on a greedy pruning algorithm for efficient inference.

\subsection{Structural max-margin learning} \label{ssec:learning}
The scoring function of Equ. \eqref{equ:score_func} is a generalized linear model w.r.t. its parameters.
Actually, we can concatenate all the model parameters to be a single vector $\mathbf{W}$ and rewrite Equ.\ \eqref{equ:score_func} by $F(\mathbf{Y}, \mathbf{Z} | \mathcal{L}, \tilde{\mathcal{S}}, \hua{H}) = \mathbf{W}^{\text{T}} \mathbf{\Phi}(\mathcal{L}, \tilde{\mathcal{S}}, \hua{H}, \mathbf{Y}, \mathbf{Z})$.
$\mathbf{\Phi}(\mathcal{L}, \tilde{\mathcal{S}}, \hua{H}, \mathbf{Y}, \mathbf{Z})$ is a re-organized sparse vector gathering all the features based on the structural state variable $(\mathbf{Y}, \mathbf{Z})$.
In our AOG model, $\mathbf{Z}$ determines the topological structure of a feasible solution (i.e., parse tree), and $\mathbf{Y}$ specifies the segments selected for the vertices of this parse tree.
Given a set of labelled examples $\{(\mathbf{Y}_n, \mathbf{Z}_n) \, | \, n = 1,2,\cdots,J\}$, we formulate a structural max-margin learning problem on $\mathbf{W}$, as shown in Equ. \eqref{equ:learning}.
\begin{eqnarray} \label{equ:learning}
& \min\limits_{\mathbf{W}} \frac{1}{2} \mathbf{W}^{\text{T}} \mathbf{W} + C \sum\limits_{n=1}^{J} \xiup_{n} \\
& \mathbf{W}^{T} \mathbf{\Phi}(\mathcal{L}_n, \tilde{\mathcal{S}}_n, \hua{H}_n, \mathbf{Y}_n, \mathbf{Z}_n) - \mathbf{W}^{T} \mathbf{\Phi}(\mathcal{L}_n, \tilde{\mathcal{S}}_n, \hua{H}_n, \mathbf{Y}, \mathbf{Z}) \nonumber \\
& \geq \Delta(\mathbf{Y}_n, \mathbf{Z}_n, \mathbf{Y}, \mathbf{Z}) - \xiup_{n}, \;\, s.t. \; \forall \,\, \mathbf{Y} \,\, \text{and} \,\, \mathbf{Z}, \nonumber
\end{eqnarray}
where $\Delta(\mathbf{Y}_n, \mathbf{Z}_n, \mathbf{Y}, \mathbf{Z})$ is a structural loss function to penalize a hypothesized parse tree $(\mathbf{Y}, \mathbf{Z})$ different from ground truth annotation $(\mathbf{Y}_n, \mathbf{Z}_n)$.
Similar to \cite{Yadollahpour13_RelativeLoss}, we adopt a relative loss as in Equ. \eqref{equ:loss}, i.e., the loss of hypothesized parse tree relative to the best one $(\mathbf{Y}^{*}, \mathbf{Z}^{*})$ that could be found from the candidate pool. That is
\begin{equation} \label{equ:loss}
\Delta(\mathbf{Y}_n, \mathbf{Z}_n, \mathbf{Y}, \mathbf{Z}) = \delta(\mathbf{Y}_n, \mathbf{Z}_n, \mathbf{Y}, \mathbf{Z}) - \delta(\mathbf{Y}_n, \mathbf{Z}_n, \mathbf{Y}^{*}, \mathbf{Z}^{*}),
\end{equation}
where $\delta(\mathbf{Y}, \mathbf{Z}, \mathbf{Y}^{'}, \mathbf{Z}^{'}) = \sum_{p \in \mathcal{T}} IoU(s^{p,z_p}_{y_p}, s^{p,z^{'}_p}_{y^{'}_p})$ is a function of measuring the part segmentation difference between any two parse trees $(\mathbf{Y}, \mathbf{Z})$ and $(\mathbf{Y}^{'}, \mathbf{Z}^{'})$.
In this paper, we employ a commonly-used cutting plane algorithm \cite{joachims2009cutting} to solve this structural max-margin optimization problem of Equ. \eqref{equ:learning}.

\subsection{Greedy Pruning Inference Algorithm} \label{ssec:inference}
On the inference of AOG models, dynamic programming or inside-outside algorithm is commonly used in literature \cite{Siskind07_Inside_Outside,Zhu12wacv_Tangram}.
However, the existence of side-way edges makes many loopy cliques in our AOG model, and thus prohibits the use of dynamic programming algorithm to efficiently infer the global optimum.
In this paper, we adopt a modified dynamic programming with greedy pruning algorithm for model inference.
This algorithm has a bottom-up scoring step followed by a top-down backtracking step.

In the first step, we recursively calculate the vertice's scores of our AOG in a bottom-up manner.
For each part $p \in \mathcal{T}$, we compute the score on every combination of its candidate segment proposal and part type $(y_{p}, z_{p})$ according to Equ. \eqref{equ:leaf_score}, and retain a top-$k$ list of the highest scored configuration values of $(y_{p}, z_{p})$.
For each part composition $c \in \mathcal{N}$, we compute the scores of subgraph rooted by vertex $c$ for candidate state configuration values based on Equ. \eqref{equ:nonleaf_score}, and also retain the top-$k$ scored configuration values of $(y_c, z_c, \{ (y_{\mu}, z_{\mu}): \mu \in Ch(c, z_c) \})$.
Particularly, only the top-$k$ configuration values of $(y_{\mu}, z_{\mu})$ are considered in further calculation for each child vertex $\mu \in Ch(c,z_c)$.
Thus the computational complexity for vertex $c$ is proportional to the number of candidate state configuration values (i. e., $\sum_{z_c=1}^{K_c} k^{|Ch(c,z_c)|}$), where $|Ch(c,z_c)|$ denotes the number of child And-nodes linked to the Or-node for state $z_c$.
In this paper, we set $k = 10$ and $|Ch(c,z_c)| \leq 3$, allowing the inference procedure tractable with a moderate quantity of state configuration values for each vertex.
It does not harm the performance notably in practice, although this greedy pruning operation gets rid of a large fraction of candidate state configuration values for part composition vertice (especially for the high-level ones, e.g. upper-body, lower-body).
We validate this issue via a diagnostic experiment in Sec. \ref{sec:exp}.

After finishing the scoring of root vertex (i.e., whole human body) in the first step, we trigger another step to backtrack the optimum state value from the retained top-$k$ list for each vertex in a top-down manner.
Concretely, for each part composition vertex $c$ we select the best scored state configuration value of
$(y_c, z_c, \{ (y_{\mu}, z_{\mu}): \mu \in Ch(c, z_c) \})$, and recursively infer the optimum state values of the selected child vertice given each $\mu \in Ch(c, z_c)$ as the root vertex of a subgraph.
In the end, we can obtain the best parse tree from the pruned solution space of our AOG, and thus output corresponding state values of $(\mathbf{Y}, \mathbf{Z})$ to produce the final parsing result.

   \vspace{-0.5\baselineskip}
\section{Experiments} \label{sec:exp}
   \vspace{-0.5\baselineskip}
In this section, we describe the details of our algorithm and conduct various experiments to demonstrate the effectiveness of our pose information in each stage of our pipeline.

\subsection{Implementation detail}
\label{ssec:details}
In part proposal selection (Sec.~\ref{ssec:part_ranking}), we train linear SVR models for $P=10$ part categories and select top $n_p=10$ segments for each part category, as candidates of the final assembling stage. We treat left part and right part as two different part categories.
Although the candidate segment pool could be type-dependent for a part/part composition in the AOG, we use common segment proposals in practice, which can significantly facilitate to compute the geometric compatibility features of side-way segment pairs.

For the segment feature in Sec.~\ref{ssec:part_assemble}, we first normalize each kind of feature independently, then concatenate them together and normalize the whole feature. All the normalization is done with $L2$ norm.
For simplicity, we only train one SVR model $g^{p}(s_i | \mathcal{L}, \hua{H})$ for each part category $p$ in Sec.~\ref{ssec:part_ranking} so that $g^{p}_{z_p} = g^{p} , \forall z_p \neq 0$ in Equ.~\eqref{equ:leaf_score}. However, due to the weight parameter $w^{p}_{z_p}$ is learned dependent on the part type $z_p$ in training the AOG, the unary terms of different part types are type-specific in the AOG model.

\subsection{Datasets and investigations} \label{ssec:exp_dataset}
\paragraph{Data.} We evaluate our algorithm on the Penn-Fudan benchmark \cite{Wang07_FudanPenn}, which consists of pedestrians in outdoor scenes with much pose variation. Because this dataset only provides testing data, following previous works \cite{Bo11_ShapeBased_PedParsing, Rauschert12_SimultShapeSeg, Luo13_PedParsing_deepDecompNet}, we train our parsing models using the HumanEva dataset \cite{Sigal06_HumanEva}, which contains 937 images with pixel-level label maps for parts annotated by \cite{Bo11_ShapeBased_PedParsing}. The labels of two datasets are consistent, which include 7 body parts: $\{$ hair, face, upper-clothes, lower-clothes, arms (arm skin), legs (leg skin), and shoes $\}$. For the pose model, we use the model provided by~\cite{Chen14_IDPR}, trained on the Leeds Sports Pose Dataset \cite{Johnson10}.

\paragraph{Effectiveness of pose for part proposal generation.}
We first investigate how our pose information help the initial part proposal generation. Specifically, we compare our pose-guided segment proposal method with the baseline proposal algorithm, i.e. the RIGOR algorithm~\cite{Humayun14_rigor} which is a faster substitute of the CPMC proposal~\cite{Carreira12_CPMC} typically used by previous parsing approaches~\cite{Dong14_UnifiedParsingPoseEst}.

For evaluating the proposal algorithms, two standard criteria are used, i.e. average part recall (APR) and average part oracle IoU (AOI). The first measures how much portion of the ground truth segments is covered by the proposals, and the second measures the best IoU we can achieve given the proposals. Formally, the APR and AOI can be written as follows,
\begin{align}
   \vspace{-0.5\baselineskip}
APR = \frac{1}{N}\sum_{i =1}^{N}{\frac{|\hua{S}_i\cap\hua{G}_i|}{|\hua{G}|}},
AOI = \frac{1}{N}\sum_{i =1}^{N}{\frac{\sum\limits_{j\in \hua{G}_i}\max\limits_{m\in\hua{S}_i}{IoU(g_j,s_m)}}{|\hua{G}_i|}},\nonumber
\end{align}
where $N$ is the number of testing images. $\hua{S}_i$ and $\hua{G}_i$ are the set of segment proposals and the set of part segment ground truth in image $i$ respectively. For computing $\hua{S}_i\cap\hua{G}_i$, we regard two segments as the same if their IoU is above 0.5.

\begin{figure}[t]
\begin{center}
   \includegraphics[width=1.02\linewidth]{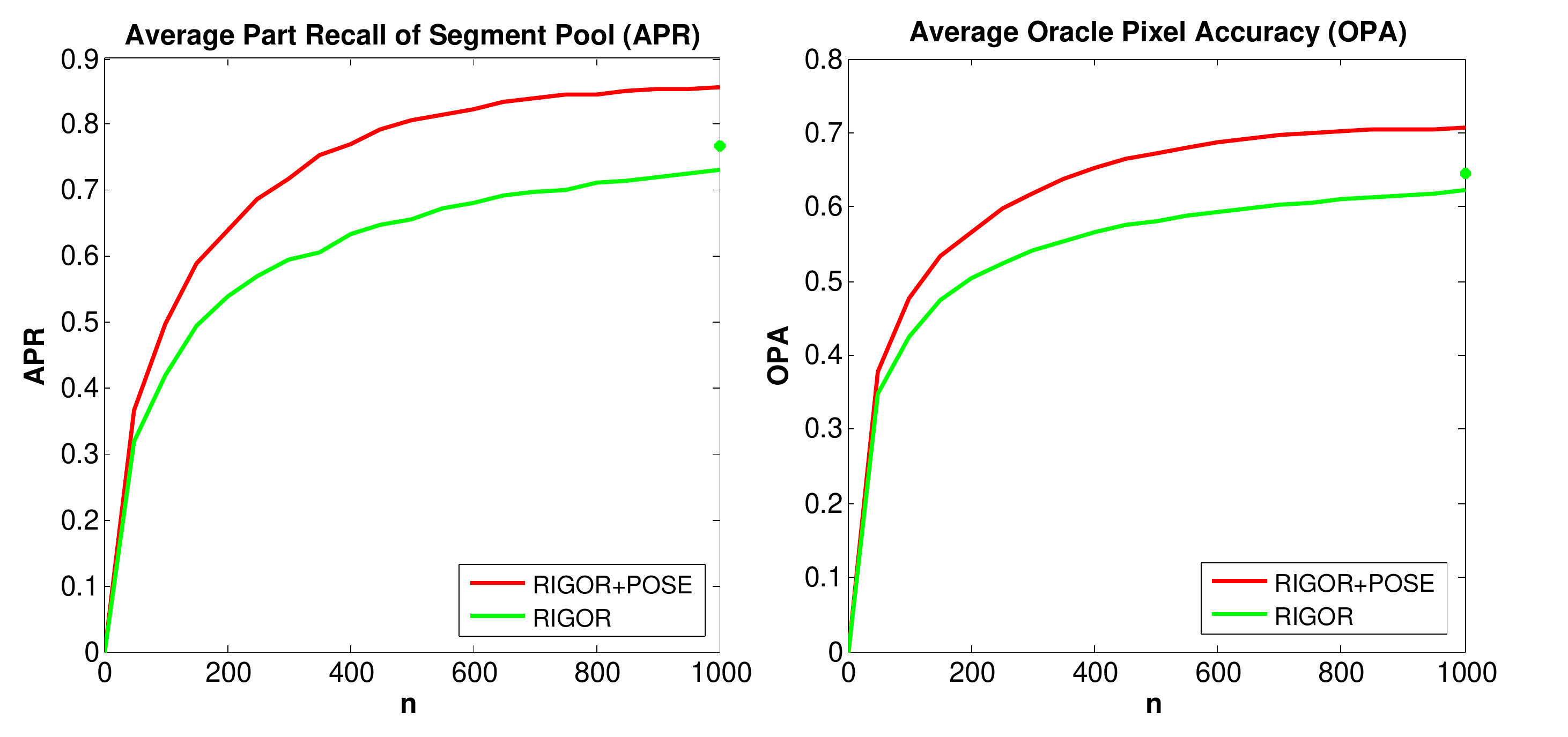}
\end{center}
   \caption{ Comparison of our part segment proposal (Red curve) to RIGOR~\cite{Humayun14_rigor} (Green curve) on human parsing over two criteria w.r.t. the number of proposals. The green asterisks on the plots represent the APR/AOI of the RIGOR pool when the pool size $n = 2000$.
}
\label{fig:result_pool}
\end{figure}

In Fig.~\ref{fig:result_pool}, we show the evaluated results over the Penn-Fudan test data. Specifically, we plot the APR and AOI w.r.t. the number of proposals upto 2000 segments. As shown in the figure, comparing with the RIGOR algorithm, ours (RIGOR + POSE) significantly improves the quality of part segment proposal by over 10$\%$ in average, which contributes much to our final performance. Finally, for each image, we select around 800 non-similar segments from the 2000 proposals as stated in Sec.~\ref{ssec:seg_prop}. In Tab.~\ref{table:pool}, we list the APR and AOI score of the segment pool composed of the selected 800 segments, capable of achieving the accuracy as high as that of the original 2000 proposals.

\begin{table}[b]
\vspace{-1\baselineskip}
\small
\centering
\setlength\tabcolsep{3pt}
\resizebox{1\columnwidth}{!}{
\begin{tabular}{c | c c c c c c c | c}
\toprule[0.2 em]
 & hair & face & u-cloth & l-cloth & arms & legs & shoes & mean\\ \midrule \midrule
 Recall & 0.88 & 0.90 & 0.99 & 0.99 & 0.86 & 0.87 & 0.67 & 0.88\\
 IoU & 0.73 & 0.74 & 0.85 & 0.86 & 0.67 & 0.72 & 0.56 & 0.73\\
\bottomrule[0.1 em]
\end{tabular}
}
\caption {Part recall and average oracle IoU of our segment pool.}
\label{table:pool}
\end{table}

\paragraph{Effectiveness of features for part proposal selection.}
To investigate various features proposed in Sec.~\ref{ssec:part_ranking} and their complementary properties, we sequentially add features into our model and show the performance of the selected part segments.
Specifically, the feature combinations we tested includes:
(1) O2P + skin;
(2) O2P + skin + PBG;
(3) O2P + skin + PBG + C-PBG;
(4) O2P + skin + PBG + C-PBG + deep potential,
which we call Model 1 to Model 4 respectively.

\begin{table}[!htpb]
\centering
\small
\setlength\tabcolsep{3pt}
\resizebox{1\columnwidth}{!}{
\begin{tabular}{c | c c c c c c c |c}
\toprule[0.2em]
Methods & hair & face & u-cloth & l-cloth & arms & legs & shoes & mean\\ \midrule \midrule
\multirow{2}{*}{o2p+skin} & 57.1 & 53.5 & 70.9 & 70.9 & 26.6 & 20.4 & 15.6 & 45.0\\
 & 68.8 & 66.9 & 80.0 & 81.4 & 54.6 & 55.3 & 45.3 & 64.6\\
 \midrule
\multirow{2}{*}{(1)+PBG} & 61.7 & 58.6 & 73.2 & 72.7 & 29.9 & 23.4 & 17.5 & 48.1\\
 & 69.9 & 66.4 & 80.6 & 82.3 & 56.4 & 54.3 & 45.8 & 65.1\\
 \midrule
\multirow{2}{*}{(2)+C-PBG} & 61.8 & 58.9 & 73.2 & 71.9 & 39.8 & 44.8 & 26.5 & 53.8\\
 & 69.9 & 66.4 & 80.5 & 82.4 & 55.8 & 59.1 & 47.4 & 65.9\\
 \midrule
\multirow{2}{*}{(3)+deep potential} & 64.4 & 59.0 & 77.4 & 77.1 & 41.4 & 43.6 & 35.1 & 56.9\\
 & 70.7 & 66.6 & 82.2 & 83.4 & 55.9 & 59.3 & 48.8 & 66.7\\
\bottomrule[0.1 em]
\end{tabular}
}
\caption{Comparison of four part models by AOI score (\%) for top-1 ranked segment (top) and top-10 ranked segments (bottom). Models are numbered as (1) to (4), from top to bottom.}
\label{table:partmodel}
\end{table}

The results are shown in Tab.~\ref{table:partmodel}, where we report the AOI score using the set of top-1 ranked part segment for each part class and top-10 ranked part segments. Firstly, we can see the results are sequentially improved, which demonstrate the effectiveness of all features we proposed. By comparing (2) and (3), we can see a significant boost of the top-1 accuracy, which indicates that after we coded the pose feature, the pose information becomes much more effective in the model. Finally, by adding the deep potential in (4), the performance of selected part segment is further improved.

We hence adopt (4) as our part ranking model to rank and select part proposals, yielding top $n_p$ ranked candidates per part category for the final assembling. The quality of selected part proposals is evaluated in Tab.~\ref{table:small_pool}. We set $n_p=10$ because it strikes a good balance between recall and segment pool size, and the oracle assembling result shows that we could surpass the state-of-the-art by over 15\% by using only the selected part segment proposals.

\begin{table}[!htpb]
\vspace{-1\baselineskip}
\centering
\small
\setlength\tabcolsep{3pt}
\resizebox{1\columnwidth}{!}{
\begin{tabular}{c | c c c c c c c | c}
\toprule[0.2em]
 & hair & face & u-cloth & l-cloth & arms & legs & shoes & mean\\ \midrule \midrule
 Recall & 0.86 & 0.79 & 0.99 & 0.99 & 0.68 & 0.69 & 0.55 & 0.79\\
 IoU & 0.71 & 0.67 & 0.82 & 0.83 & 0.56 & 0.59 & 0.49 & 0.67\\
Oracle & 0.73 & 0.70 & 0.83 & 0.84 & 0.62 & 0.64 & 0.45 & 0.69\\
\bottomrule[0.1 em]
\end{tabular}
}
\caption {Evaluation of our selected segment pool (top-10 segments per category) in terms of part recall, AOI score, and oracle assembling pixel accuracy.}
\label{table:small_pool}
\end{table}

\paragraph{Effectiveness of the AOG.}
To show the effectiveness of our AOG design, we set up two experimental baselines for comparison. (1) Naive assembling: consider only the unary terms and basic geometric constraints as defined in~\cite{Bo11_ShapeBased_PedParsing}, e.g. upper-clothes and lower-clothes must be adjacent.
(2) Basic AOG: consider only the unary terms and the vertical edges, without pairwise side-way edges from the C-PBG feature in Eqn.~(\ref{equ:leaf_score}).
We show the results in Tab.~\ref{tab:aog_model}, where the Basic AOG with vertical relations outperforms simple hard constraints, and by adding the pairwise side-way edges, the results are further boosted, which strongly supports each component of our AOG model. We also display the results for our model without pruning for comparison, which clearly justifies our use of pruning in AOG inference. We can see that pruning only brings ignoreable decrease in performance while it reduces the inference time from 2 min. to 1 sec. per image.

\begin{table}[t]
\small
\centering
\setlength\tabcolsep{3pt}
\resizebox{1\columnwidth}{!}{
\begin{tabular}{c | c c c c c c | c }
\toprule[0.2 em]
Methods & hair & face & u-cloth & arms & l-cloth & legs & Avg \\   \midrule \midrule
Naive Assembling & 62.3 & 53.5 & 77.8 & 36.9 & 78.3 & 28.2 & 56.2 \\
Basic AOG & 63.1 & 52.9 & 77.1 & 38.0 & 78.1 & 35.9 & 57.5 \\
Ours & \textbf{63.2} & \textbf{56.2} & \textbf{78.1} & \textbf{40.1} & \textbf{80.0} & 45.5 & \textbf{60.5} \\
Ours (w/o pruning) & 63.2 & 56.2 & 78.1 & 40.1 & 80.0 & \textbf{45.8} & 60.5 \\
\bottomrule[0.1 em]
\end{tabular}
}
\caption {Per-pixel accuracy (\%) of our AOG and two baselines.}
\label{tab:aog_model}
\vspace{-1\baselineskip}
\end{table}

\paragraph{Training and testing time.}
Generally, for training, our approach takes two days for training a FCN model and two day for training the SVR selection model and AOG model. For testing, it takes 6s for extracting various features and around 1s to assemble the segments using AOG with our MATLAB implementation.




%

\subsection{Comparison to the state-of-the-art} \label{sssec:exp_perf}
We compare our approach with four state-of-the-art methods, namely FCN~\cite{long_shelhamer_fcn}, SBP \cite{Bo11_ShapeBased_PedParsing}, P\&S \cite{Rauschert12_SimultShapeSeg}, and DDN \cite{Luo13_PedParsing_deepDecompNet}, which use the same training and testing settings.
Specially, for FCN, we use the code provided by the author and re-train a model with our training set. Following~\cite{peng_part}, we train the FCN up to 16s version since there is little improvement using the 8s version. We refer the reader to~\cite{long_shelhamer_fcn, peng_part} for more details due to the limited space.


\begin{table}[!htpb]
\vspace{-1\baselineskip}
\centering
\small
\setlength\tabcolsep{3.5pt}
\resizebox{1\columnwidth}{!}{
\begin{tabular}{c | c c c c c c c | c}
\toprule[0.2 em]
Method & hair & face & u-cloth & arms & l-cloth & legs & shoes & Avg$^{*}$ \\ \midrule \midrule
FCN-32~\cite{long_shelhamer_fcn} & 50.2 & 33.7 & 69.4 & 13.8 & 66.7 & 14.2 & 25.2 &41.3 \\ 
FCN-16~\cite{long_shelhamer_fcn} & 48.7 & 49.1 & 70.2 & 33.9 & 69.6 & 29.9 & \textbf{36.1} & 50.2 \\ 
P\&S~\cite{Rauschert12_SimultShapeSeg} & 40.0 & 42.8 & 75.2 & 24.7 & 73.0 & 46.6 &- & 50.4 \\
SBP~\cite{Bo11_ShapeBased_PedParsing} & 44.9 & \textbf{60.8} & 74.8 & 26.2 & 71.2 &42.0 &- &  53.3 \\ 
DDN~\cite{Luo13_PedParsing_deepDecompNet} & 43.2 & 57.1 & 77.5 & 27.4 & 75.3 & \textbf{52.3} &- & 56.2 \\
Ours & \textbf{63.2} & 56.2 & \textbf{78.1} & \textbf{40.1} & \textbf{80.0} & {45.5} & 35.0 & \textbf{60.5} \\
\bottomrule[0.1 em]
\end{tabular}
}
\caption{Comparison of our approach with other state-of-the-art algorithms over the Penn-Fudan dataset. The Avg$^{*}$ means the average without shoes class since it was not included in other algorithms.}
\label{table:res_8}

\end{table}
\begin{figure}[t]
\begin{center}
\begin{tabular}{c}
\includegraphics[width=0.9\linewidth]{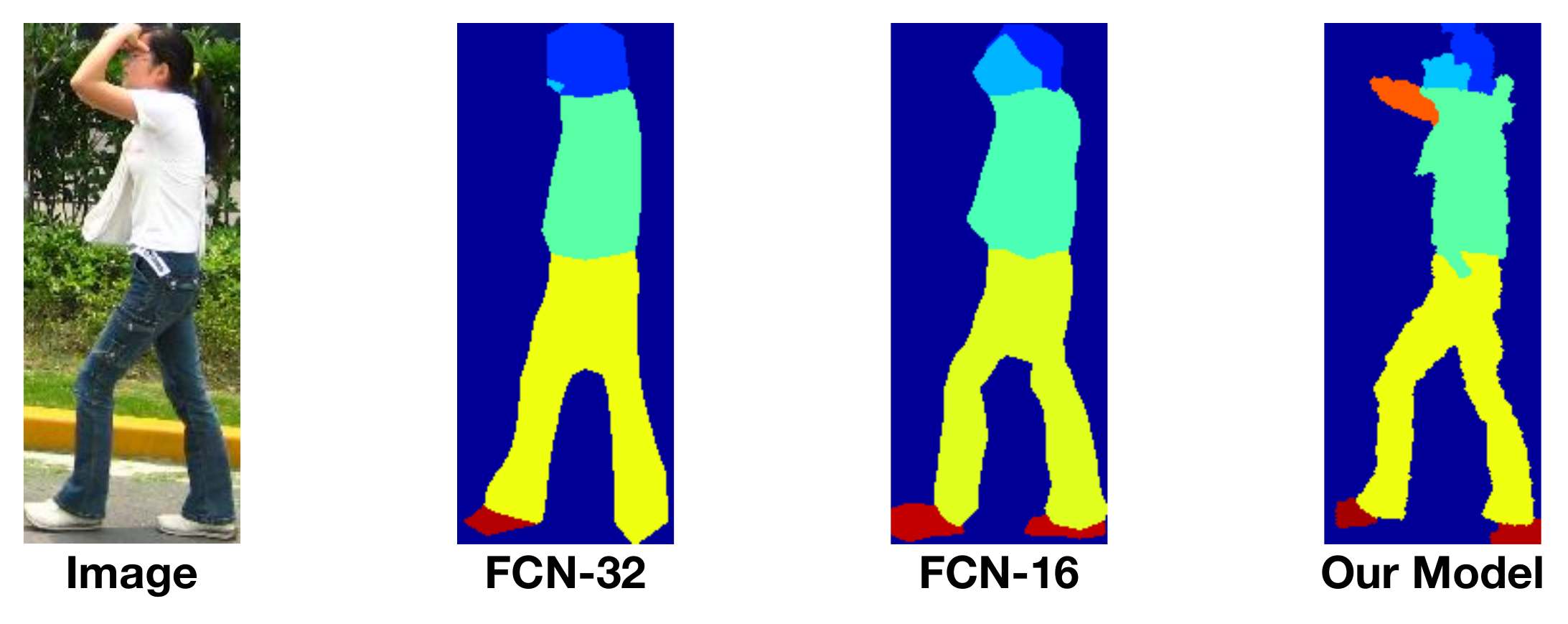}\\
(a) Visual comparison between ours and FCN~\cite{peng_part}\\
\includegraphics[width=0.95\linewidth]{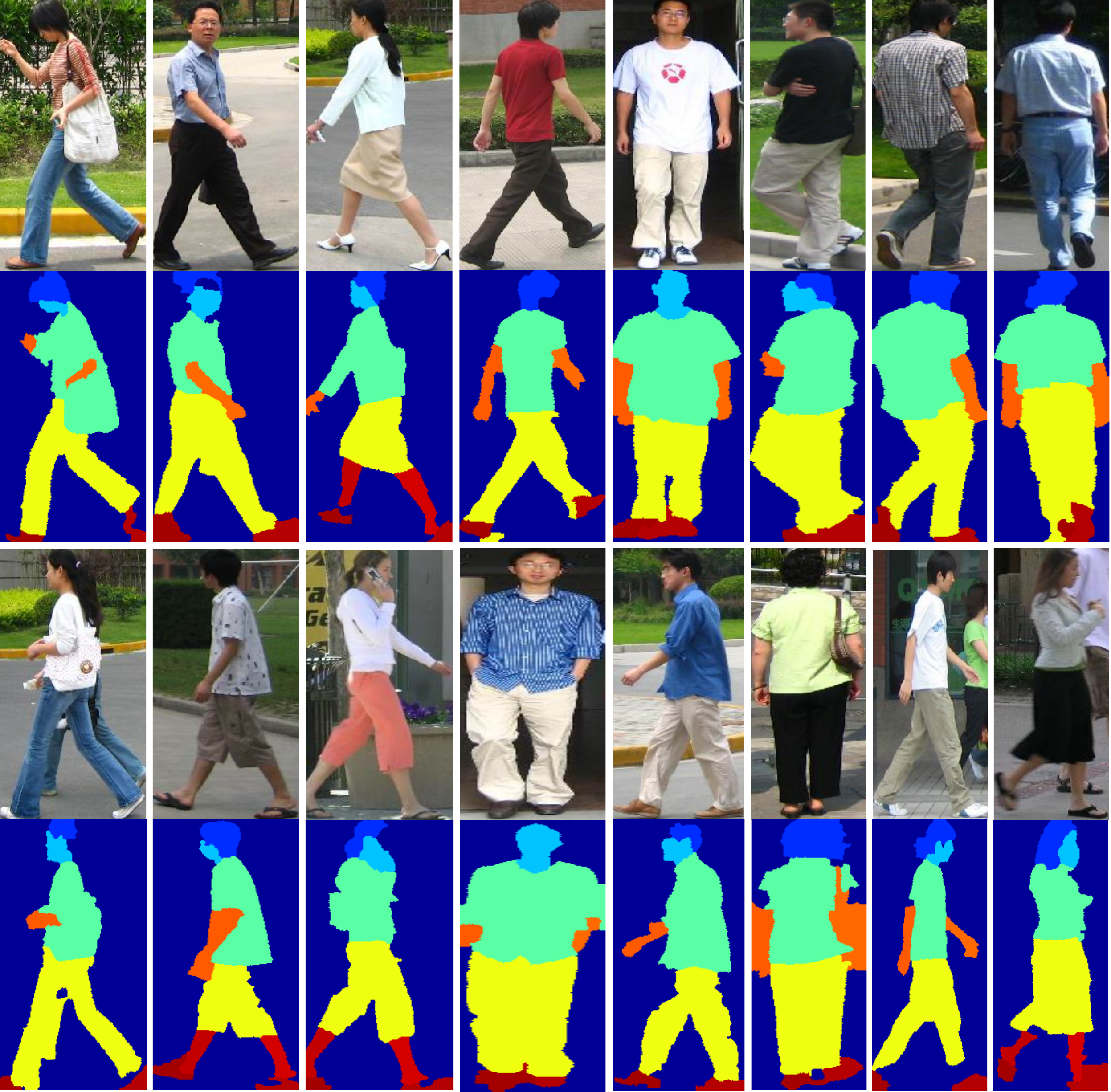}\\
(b) Additional parsing results.
\end{tabular}
\end{center}
  \caption{Qualitative results of our algorithm over the Penn-Fudan dataset~\cite{Wang07_FudanPenn}. }
  \vspace{-1\baselineskip}
\label{fig:result5}
\end{figure}

\begin{figure}[b]
\begin{center}
   \includegraphics[width=1.00\linewidth]{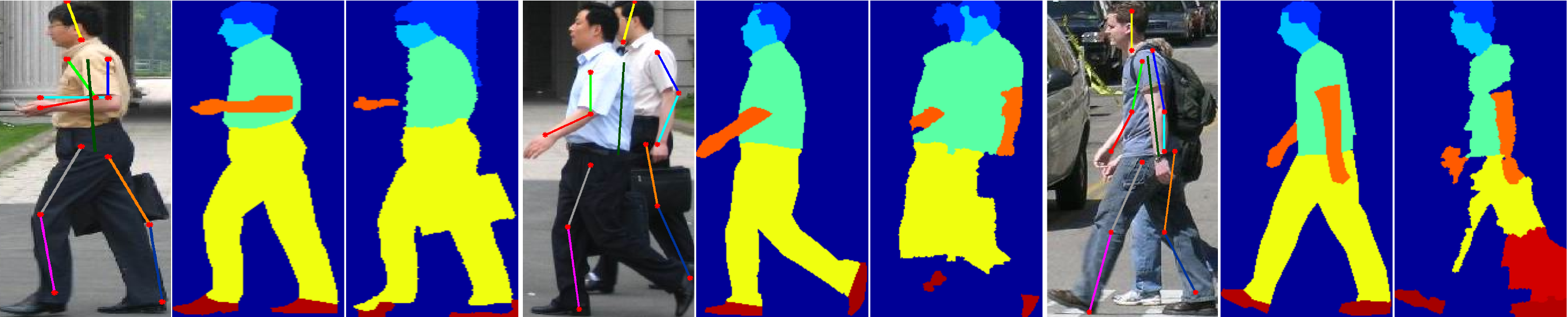}
\end{center}
   \caption{Failure cases of our algorithm on Penn-Fudan dataset. For each case, the original image (with pose prediction), ground truth, and our parsing result are displayed from left to right.}
\label{fig:result6}
\end{figure}

\paragraph{Quantitative results.}
We show the comparison results in Tab.~\ref{table:res_8}. Our model outperforms the potential directly from FCN by over 10$\%$ and the state-of-the-art DDN model by over 4$\%$, from which we can see most improvement is from the small part such as hair and arms. 
It demonstrates that our pose information produces strong segment candidates that align to boundaries for small parts. In addition, our pose feature design together with deep potentials and AOG models obtains long-range context information, which gives our model robustness in shape variation and avoid many local confusions.

\paragraph{Qualitative results.}  As the authors of other comparisons did not release their code and visual results, qualitatively, we can only compare our results with the FCN as shown in Fig.~\ref{fig:result5}(a). From the results, FCN can produce high accuracy over the general shape of the human while missing align to some local details like arms and shoes. Ours solve such problems with pose-guided local segment proposal, robust segment selection and assembling, yielding more satisfied results.

In Fig.~\ref{fig:result6}, we show three failed examples due to color confusion with other objects, multiple instance occlusion, and big variation in lighting respectively, which generally fail most of the curent algorithms. For the first case and the third case, we got accurate pose prediction but failed to generate satisfying segment proposals for lower-clothes, which suggests that we either adopt stronger shape cues in the segment proposal stage or seek richer context information (e.g. handbag in the first case). For the second case, we got a bad pose prediction in the beginning and thus mixed two people's parts during assembling, which indicates the necessity of handling instance pose estimation or segmentation, which is beyond the scope of this paper.

   \vspace{-0.5\baselineskip}
\section{Conclusion} \label{sec:conclusion}
   \vspace{-0.5\baselineskip}
In this paper, we present a human parsing pipeline which integrates human pose information and deep-learned features into segments, producing robust human parsing results.  Our approach majorly includes three stages: part segment proposal, part proposal selection, and part assembling with an And-Or graph. In this framework, we systematically explore human pose information, including the pose-guided proposal and novel pose-based features which successfully applied to the part selection and assembling.
We did extensive experiments that validate the effectiveness of each components of our framework, and finally, our approach significantly surpasses other state-of-the-art methods on the popular Penn-Fudan benchmark~\cite{Wang07_FudanPenn}.

The future work includes several aspects: (1) Adopt useful shape cues for the part proposal and selection stages. (2) Use the learned part prototypes to define part sub-types and explicitely incorporate them into the AOG model; (3) Combine CNN with graphical models in a more efficient way to better utilize their complementary role in the human parsing task.

{\small
\bibliographystyle{ieee}
\bibliography{egbib_v1.2}
}

\end{document}